\documentclass[manuscript,authorversion,noacm]{acmart}
\AtBeginDocument{%
  \providecommand\BibTeX{{%
    \normalfont B\kern-0.5em{\scshape i\kern-0.25em b}\kern-0.8em\TeX}}}

\setcopyright{acmcopyright}

\usepackage{cleveref}

\copyrightyear{2023} 
\acmYear{2023} 
\setcopyright{acmlicensed}\acmConference[IUI '23]{28th International Conference on Intelligent User Interfaces}{March 27--31, 2023}{Sydney, NSW, Australia}
\acmBooktitle{28th International Conference on Intelligent User Interfaces (IUI '23), March 27--31, 2023, Sydney, NSW, Australia}
\acmPrice{15.00}
\acmDOI{10.1145/3581641.3584066}
\acmISBN{979-8-4007-0106-1/23/03}

\begin{document}

\title{Appropriate Reliance on AI Advice: Conceptualization and the Effect of Explanations} 

\author{Max Schemmer}
\affiliation{%
  \institution{Karlsruhe Institute of Technology}
  \city{Karlsruhe}
  \country{Germany}
  }
\email{max.schemmer@kit.edu}

\author{Niklas Kühl}
\affiliation{%
  \institution{Karlsruhe Institute of Technology}
  \city{Karlsruhe}
  \country{Germany}
  }
\email{niklas.kuehl@kit.edu}

\author{Carina Benz}
\affiliation{%
  \institution{Karlsruhe Institute of Technology}
  \city{Karlsruhe}
  \state{Baden-Württemberg}
  \country{Germany}
  }
\email{carina.benz@.kit.edu}

\author{Andrea Bartos}
\affiliation{%
  \institution{Karlsruhe Institute of Technology}
  \city{Karlsruhe}
  \state{Baden-Württemberg}
  \country{Germany}
  }
\email{andrea.bartos@alumni.kit.edu}

\author{Gerhard Satzger}
\affiliation{%
  \institution{Karlsruhe Institute of Technology}
  \city{Karlsruhe}
  \state{Baden-Württemberg}
  \country{Germany}
  }
\email{gerhard.satzger@.kit.edu}




\begin{abstract}
AI advice is becoming increasingly popular, e.g., in investment and medical treatment decisions. As this advice is typically imperfect, decision-makers have to exert discretion as to whether actually follow that advice: they have to “appropriately” rely on correct and turn down incorrect advice.
However, current research on appropriate reliance still lacks a common definition as well as an operational measurement concept. Additionally, no in-depth behavioral experiments have been conducted that help understand the factors influencing this behavior.
In this paper, we propose Appropriateness of Reliance (AoR) as an underlying, quantifiable two-dimensional measurement concept. We develop a research model that analyzes the effect of providing explanations for AI advice. 
In an experiment with 200 participants, we demonstrate how these explanations influence the AoR, and, thus, the effectiveness of AI advice.
Our work contributes fundamental concepts for the analysis of reliance behavior and the purposeful design of AI advisors.

\end{abstract}

\ccsdesc[500]{Human-centered computing~Empirical studies in HCI}
\ccsdesc[500]{Computing methodologies~Artificial intelligence}

\keywords{Appropriate Reliance, Explainable AI, Human-AI Collaboration, Human-AI Complementarity 
}


\maketitle

\section{Introduction}
Most important decisions are made by calling upon advisors.
While in the past advice was typically obtained from human experts, nowadays advisors based on artificial intelligence (AI) are becoming increasingly frequent in research and practice \cite{jung2018robo,jussupow2021augmenting}. For example, AI advises medical professionals in breast cancer screening \cite{mckinney2020international}, or in loan decisions\cite{demajo2020explainable}. 

In the past, research has predominantly focused on maximizing the reliance \cite{kerasidou2021before}, trust \cite{siau2018building}, utilization \cite{alnowami2022regression}, compliance \cite{kuhl2020you}, or acceptance \cite{shin2021effects} of AI advice \cite{jussupow2021augmenting}. With certain nuances, all these terms basically describe a concept that aims at maximizing the amount of AI advice that a human decision-maker eventually follows, i.e. maximizing the AI reliance. 
Recently, a new line of research has emerged that argues that maximising AI reliance does not fully exploit the potential of state-of-the-art human-AI decision making \cite{bansal2021does,zhang2020effect,buccinca2021trust,schemmer2022should}. We summarize the reasons for this line of thought in three main categories:

\textbf{\textit{Increasing usage of imperfect AI advisors. }} First, prior research on AI advice often assumed  ``perfect'' advice \cite{jussupow2021augmenting}. This makes sense if one considers narrow application spaces, e.g. performing deterministic algebra. But AI nowadays is used for more complex tasks \cite{frey2017future} which increases both the number and severity of AI errors. Therefore, generally accepting AI advice would also include the acceptance of incorrect AI advice: 
Assume that without an AI advisor's intervention, a physician would have diagnosed cancer, yet in the setting with AI advice had been misled by an incorrect AI advice and, thus, had failed to detect the cancer. 

\textbf{\textit{Increasing alignment of the objectives of human decision-makers and AI advisors. }} 
Second, in previous research, the goal of human or AI advisors was often inconsistent with the goal of the decision-maker, e.g., financial advisors wanted to persuade a client to purchase certain financial products that would yield the highest financial benefit for their own bonus. In this situation, AI designers do not want customers to differentiate good from bad advice but simply increase acceptance of advice.
Today, however, AI advisors are often specifically designed to enhance human decision-making \cite{lai2021towards}. The advice seekers can design the advisor based on individual goals and with desired features such as ``honest'' explainability. This honesty might be missing if the inherent goal of the advisor is different from the advice seeker's.
If the AI is not designed in alignment with human goals, it is often not in the AI designer's interest to enable humans to critically question the advice they receive. 

\textbf{\textit{Increasing potential for complementary team performance. }}
Third, as modern AI is not only more performant and offers more application areas, but also complements humans \cite{hemmer2022effect,lai2022human,hemmer2022forming}, there is an increasing potential to achieve complementary team performance (CTP), i.e., a performance that exceeds both---individual AI and human performance \cite{bansal2021does,hemmer2022effect,fugener2021will,nguyen2022visual,lai2022human}. However, this level of performance can only be achieved by exploiting complementary capabilities. Even in the case of superior AI advice, it cannot be achieved simply by accepting AI advice, as it is then tied to AI performance without considering the potential additional strengths of the human.

In conclusion, human decision-makers should not simply 
rely on AI advice, but should be empowered to differentiate when to rely on AI advice and when to rely on their own, i.e. they should display \textit{appropriate reliance} (AR) \cite{bansal2021does,wang2021explanations,yang2020visual,zhang2020effect}.
Despite being a necessary condition for the effective use of AI, current research on AR on AI advice is still very ambiguous with regard to definition, measurement, and impact factors.\footnote{It is worth to mention that several studies have examined AR in automation and robotics research \cite{lee2004trust,wang2008selecting}, but an agreed-upon definition of AR and a respective metric are still missing \cite{wang2008selecting,lai2021towards}.}

First, we deal with the ambiguous concept of AR: Current research inconsistently uses the term both for a binary target state (``appropriate reliance is either achieved or not'') and a metric indicating a degree of appropriateness. To clarify this, we introduce a two-dimensional metric---the appropriateness of reliance (AoR)---to describe and measure reliance behavior. It is based on relative frequencies of correctly overriding wrong AI suggestions (
self-reliance) and following correct AI suggestions (
AI reliance), and reflects a metric understanding of AR. This metric can then be used to define different levels as target states of AR that mark the achievement of particular objectives like certain legal, ethical and performance requirements.
 
Second, we analyze how the provision of explanations influences AoR and the achievement of AR states. Existing literature is ambiguous with regard to effects of explanations \cite{bansal2021does,alufaisan2021does,wang2021explanations}: While in some experiments, explanations support AR \cite{wang2021explanations,yang2020visual},  in others they cause ``blind trust'' \cite{bansal2021does,alufaisan2021does} in AI advice. To better understand and reconcile conflicting results, we consider additional constructs that may mediate the effect of explanations. More specifically, we hypothesize that explanations do not only influence the information available to the decision-maker, but also have an impact on trust toward AI and on self-confidence.

To test our hypotheses, we conduct a behavioral experiment with 200 participants.
Our experiment underscores the advantages of AoR as a metric to examine in detail the factors that lead to changes in overall performance. 
Moreover, our results help explain the relationship between explanations and AoR by assessing the role of reliance and confidence as mediators, thus mitigating the ambiguity of previous research.

Our work provides researchers with a theoretical foundation of AR within human-AI decision-making research and provides guidance on how to design AI advisors.
More specifically, our research contributes to research and practice by
defining AR, developing a measurement concept (AoR), and analyzing how explainable AI influences the AoR. Our definition should help researchers to more accurately describe whether they have achieved AR in their experiments. The AoR metric allows to precisely steer the development towards AR. Lastly, our experimental insights can be seen as a starting point for in-depth experimental evaluations of factors impacting AoR. 

The remainder of this article is structured as follows: In \Cref{sec:RW}, we first outline related work on AR in the context of human-AI decision-making. 
In \Cref{sec:concept}, we define AR and develop a measurement concept, the AoR, to isolate different possible effects.
Subsequently, we derive impact factors on AoR in \Cref{sec:RM}. 
In \Cref{sec:ED}, we describe the design of our behavioral experiment and summarize the results in \Cref{sec:Results}. In \Cref{sec:Discussion}, we discuss our results and provide ideas for future work. 
\Cref{sec:Conclusion} 
concludes our work.

\section{Related Work}\label{sec:RW}
In the following, we introduce the related work of this article, structured along the topics of appropriate reliance in human advice, automation, and human-AI decision-making as well as the role of explainability. 

\textbf{Appropriate reliance in human advice.} Historically, the use of (human) advice is generally discussed in the so-called judge-advisor system (JAS) research stream \cite{harvey1997taking,yaniv2004receiving,sniezek2001trust}. The term ``judge'' refers to the decision-maker who receives the advice and must decide what to do with it \cite{bonaccio2006advice}. The judge is the person responsible for making the final decision. The ‘‘advisor’’ is the source of the advice \cite{bonaccio2006advice}. The research stream mainly focuses on advice acceptance.\footnote{In this article, we use the term advice acceptance as a generic term to describe the behavior of following AI advice regardless of its quality.}

\textbf{Appropriate reliance in automation.} In contrast, many researchers have worked on AR with regard to automation \cite{lee2004trust} and robotics \cite{talone2019effect}. In the following, we will provide an overview of the most common definitions. Fundamental work in the context of AR in automation has been laid by \citet{lee2004trust}. The authors outline the relationship between ``appropriate trust'' and AR in their work. However, they do not define AR explicitly but provide examples of inappropriate reliance, such as ``misuse and disuse are two examples of inappropriate reliance on automation that can compromise safety and profitability'' \cite[p. 50]{lee2004trust}.
\citet{wang2008selecting} define appropriate reliance as the impact of reliance on performance. For example, they discuss the situation in which automation reaches a reliability of 99\%, and the human performance is 50\%. In their opinion, it would be appropriate to always rely on automation as this would increase performance. \citet{talone2019effect} follows the work by \citet{wang2008selecting} and defines AR as ``the pattern of reliance behavior(s) that is most likely to result in the best human-automation team performance'' \cite[p. 13]{talone2019effect}. Both see AR as a function of human-automation team performance.

\textbf{Appropriate reliance in human-AI decision-making.}
Recent work in human-AI decision-making has started to discuss AR in the context of AI advice. \citet{lai2021towards} give an overview of empirical studies that analyze AI advice considering AR. For example, \citet{Chandrasekaran2018DoEM} analyze whether humans can learn to predict the AI's behavior. This ability is associated with an improved ability to appropriately rely on AI predictions. Moreover, \citet{gonzalez2020human} measure the acceptance of incorrect and correct AI advice. 
Similarly, \citet[p. 1]{poursabzi2021manipulating} point out the idea of AR in the form of ``making people more closely follow a model’s predictions when it is beneficial for them to do so or enabling them to detect when a model has made a mistake''. However, the authors do not explicitly relate this idea to the concept of AR. In this context, additional work uses the term ``appropriate trust'' with a similar interpretation as the behavior to follow ``the fraction of tasks where participants used the model's prediction when the model was correct and did not use the model's prediction when the model was wrong'' \cite[p. 323]{wang2021explanations}. Finally, also \citet[p. 190]{yang2020visual} define ``appropriate trust is to [not] follow an [in]correct recommendation''. All these articles have in common that they consider AR or appropriate trust on a case-by-case basis.
\citet{Bussone2015TheRO} assess how explanations impact trust and reliance on clinical decision support systems. The authors divide reliance into over- and self-reliance as part of their study. They use a qualitative approach to answer their research questions. 
\citet{chiang2021you} evaluate the impact of tutorials on AR and measure AR through team performance.

The related work highlights that current research inconsistently uses the term both for a binary target state (``AR is either achieved or not'') and a metric indicating a degree of appropriateness. Additionally, previous research does not provide a unified measurement concept that allows measuring the degree of appropriateness \cite{lai2021towards}.

\textbf{Explainable AI and appropriate reliance.} 
Most researchers that studied AR so far have proposed to use explanations of AI as a means for AR \cite{adadi2018peeking,lai2019human,zhang2020effect}. We refer to AI that generates explanations as explainable AI (XAI). 
Explanations can be differentiated in terms of their scope, i.e., being global or local explanations \cite{adadi2018peeking}. Global XAI techniques address holistic explanations of the models as a
whole. In contrast, local explanations work on an individual instance basis. Besides the scope, XAI
techniques can also be differentiated with regard to being model specific or model agnostic, i.e., whether they can
be used with all kinds of models \cite{adadi2018peeking}.
The most commonly used model agnostic technique is feature importance \cite{lundberg2017unified,lundberg2018consistent}. Feature importance can be used to generate saliency maps for computer vision tasks or highlight important words for text classification. 

\begin{table}[H]
\caption{Related work on explainable AI (XAI) and Appropriate Reliance (AR).}
\label{tab:RW}
\begin{tabular}{lllc}
\toprule
Study & AR Metric & Independent variable & Effect of XAI on AR 
\\
\midrule
\cite{bansal2021does} & Accuracy on correct or incorrect AI advice & Local feature importance  &  Negative
  
\\
 & & Adaptive explanations 
 & Positive\\
  & & based on AI confidence 
 & 
\\
\cite{buccinca2021trust} & Ratio of reliance on incorrect AI advice & Local feature importance  &  Negative

\\
\cite{jakubik2023empirical} & Ratio of reliance on correct & Local feature importance &   No effect  
\\
& or incorrect AI advice & Predictive outcomes & Negative
\\
\cite{wang2021explanations} & Accuracy on correct or incorrect AI advice & Global feature importance &  No effect
\\
 &  & Local feature importance & Positive
 \\
 &  & Examples &  No effect  
  \\
 &  & Counterfactuals &  No effect  
\\
\cite{yang2020visual} & Ratio of reliance on correct & Local feature importance &   Positive
\\
& or incorrect AI advice 
\\
\bottomrule
\end{tabular}
\end{table}

Several studies have evaluated whether different types of explanations can support humans' understanding of the AI model with the goal of better relying on recommendations in the correct cases \cite{alufaisan2021does,buccinca2021trust,Carton2020,VanderWaa2021}. 
However, it has also been shown that some types of explanations can lead people to rely too much on the AI's recommendation, especially in cases where the AI advice is wrong \cite{bansal2021does,poursabzi2021manipulating,schemmer2022influence}. 
In \Cref{tab:RW} on page \pageref{tab:RW}, we provide a comprehensive overview of the results that were found in the current studies on AR in XAI-assisted decision-making.
Overall, we find mixed results regarding the effect of explanations.

To sum it up, related work is missing a precise definition of AR, a unified measurement concept, and a precise understanding of when and why explanations of AI advisors influence AR.

\section{Conceptualization of Appropriate Reliance}\label{sec:concept}

Although several studies have examined AR in automation and robotics research \cite{lee2004trust,wang2008selecting}, an agreed-upon definition of AR and a respective metric are still missing \cite{wang2008selecting,lai2021towards}. 
We, therefore, initiate our research by deriving a definition of AR and a corresponding metric.
To do so, we first discuss the terms reliance and appropriateness. Following that, we derive our metric and lastly define AR.

\subsection{Reliance and Appropriateness}
Reliance itself is defined as a behavior \cite{lee2004trust,dzindolet2003role}. This means it is neither a feeling nor an attitude but the actual action conducted. This means reliance is directly observable. \citeauthor{scharowski2022trust} define reliance in the AI advisor context as ``user’s behavior that follows from the advice of the system'' \cite[p. 3]{scharowski2022trust}.
Defining reliance as behavior also clarifies the role of trust in this context, which is defined as ``the attitude that an agent will help achieve an individual’s goals in a situation characterized by uncertainty and vulnerability'' \cite[p. 51]{lee2004trust}. In general, research has shown that trust in AI increases reliance, but reliance can also take place without trust being present \cite{lee2004trust}. For example, we might not trust the bank advisor but consciously decide that following the advice is still the best possible decision. The other way around, we could also generally trust an advisor, but consciously decide that the given advice is not correct in a particular situation.
Finally, reliance is influenced beyond trust by other attitudes such as perceived risk or self-confidence \cite{riley2018operator}.

After establishing a common understanding of reliance, we proceed by defining ``appropriateness''. 
Appropriateness depends on different types of AI errors. Current AI is imperfect, i.e., it may provide erroneous advice. This erroneous advice can be divided into systematic errors and random errors \cite{talone2019effect}. While humans can identify systematic errors, random errors have no identifiable patterns and can not be distinguished. These different types of errors allow differentiation between two cases. If all errors are random and cannot be detected, then, from a performance perspective, humans should always rely on the AI's advice when the AI performs better on average, and never when the AI performs worse on average\cite{talone2019effect}. 
However, suppose there are some systematic errors. In that case, humans might be able to differentiate between correct and incorrect advice, which may even result in superior performance , i.e. complementary team performance (CTP) \cite{hemmer2021human}, compared to a scenario in which AI and humans conduct the task independently of each other.

This changes the overall discrimination to a case-by-case discrimination. In the presence of systematic errors, humans should evaluate each case individually. Since the solution approach in the presence of just random errors is relatively simple, as pointed out above, in this article, we focus on the more complicated setting when a significant proportion of task instances exhibit systematic errors. 
After having defined the terms reliance and appropriateness in the following, we derive a metric.

\subsection{Towards a Measurement Concept---Appropriateness of Reliance }
Appropriateness is often measured by the percentage in which the decision-maker relies on correct AI advice and the percentage in which the decision-maker does not rely on incorrect AI advice \cite{bansal2021does,gonzalez2020human}. 
The major disadvantage of this measurement is that we cannot know whether the reliance on correct AI advice stems from a correct discrimination or simply an overlap of the human and the AI's decision, i.e. instances where the AI advisor just confirms the human \cite{tejeda2022ai}. 
Especially from an ethical point of view, it makes a major difference whether the final decision is incorrect because an AI advisor ``convinced'' a human decision-maker to accept an incorrect AI advice or whether the human decision-maker would also not have been competent to solve it alone.  

Therefore, to measure the degree of appropriateness in a more narrow sense, we follow the approach of the judge-advisor system (JAS) paradigm and include an initial human decision \cite{sniezek2001trust}. This approach requires participants to make a decision, receive advice, and then make a second, potentially revised decision. In general, if we do not consider the initial human decision, information about the human discrimination ability, including the consequent action, gets lost---it is not traceable how the human would have decided without the AI advice. Nevertheless, especially this interaction needs to be documented to research AR holistically.

\begin{figure}[h]
  \centering
  \includegraphics[width=0.7\linewidth]{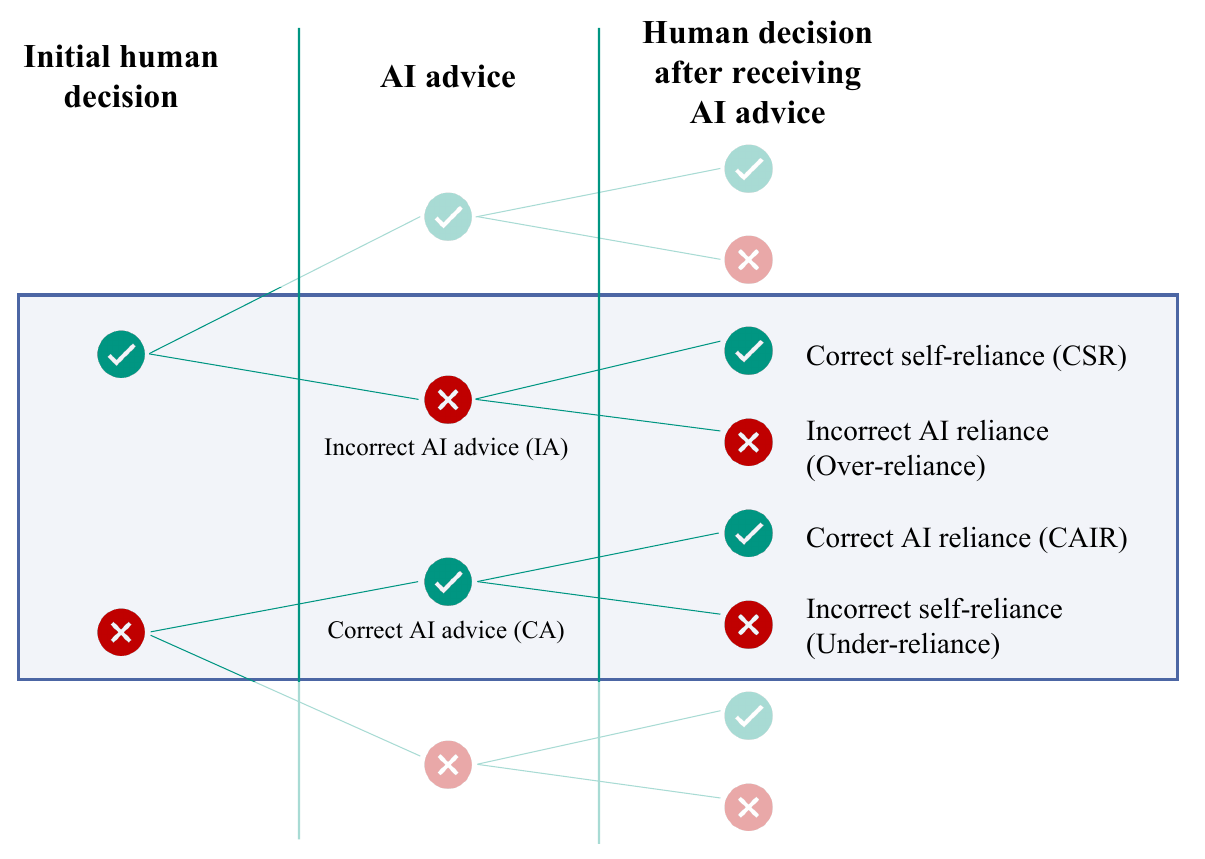}
  \caption{Combinatorics of initial human decisions, AI advice and human reliance for a single task instance in a sequential task setting.}
  \label{fig:metric}
\end{figure}

We use a simple discrete decision case to highlight the different possible outcomes of reliance. Note that for simplicity, we refer to classification problems. However, the measurement concept can be extended to regression problems as well (see for example \citet{petropoulos2016big}).
We focus on a single task instance perspective and consider a sequential decision process which can be described as follows: First, the human makes a decision, then receives AI advice. Second, the human is asked to update the initial decision, i.e., either adopt or overwrite the AI advice. 
This allows measuring appropriateness in a fine-granular way. 
\Cref{fig:metric} on page \pageref{fig:metric} highlights the different combinations.  
Four of the eight combinations are cases where the human's initial decision and the AI's advice are the same, i.e. the AI confirms the human's decision. In our reliance measure, we exclude these confirmation cases for two reasons:
First, if the same decision is made in all three steps, it is impossible to objectively measure whether AI or self-reliance was present. Second, if the final decision differs from the advice and the initial human decision, it is questionable whether we can speak of a reliance outcome. For example, if both the human and the AI are initially incorrect, then arriving at a correct final decision is less a matter of reliance than of human-AI collaboration. 
While these cases of collaboration are relevant to CTP, they are beyond the scope of our work.
Therefore, we arrive at four different reliance outcomes, which we present below.

On a high level, we can cluster those four outcomes into either AI or self-reliance.
\textit{AI reliance} refers to cases in which the initial decision-maker's decision is different from the AI advisor and the decision-maker relies on the AI's advice. Likewise, \textit{self-reliance} refers to cases in which the decision-maker is different from the AI advisor but finally relies on themselves. 
On a more detailed level, we can further differentiate whether the final decision is correct or incorrect which leaves us with the following four reliance outcomes:
First, \textit{correct AI reliance (CAIR)}, which describes the case when the human is initially incorrect, receives correct advice, and relies on that advice. Second, the case in which the human relies on the initial incorrect decision and neglects correct AI advice. This is denoted as \textit{incorrect self-reliance} or under-reliance. Third, if the human is initially correct and receives incorrect advice, this can either result in \textit{correct self-reliance (CSR)}, i.e., neglecting the incorrect AI advice, or relying on it, which is denoted as \textit{incorrect AI reliance} or over-reliance.\footnote{This also clarifies our definition of over- and under-reliance. Both are errors on a task instance level, when humans do not rely appropriately.} 
Based on these cases, we propose a two-dimensional metric that transfers the instance perspective on a measurement for multiple task instances.

Let us consider a prediction task $T = \{X_i, y_i\}_i^N$ as a set of $N$ instances $x_i \in X$ with a corresponding ground truth label $y_i \in Y$. 
On the first dimension, we calculate the ratio of the number of cases in which humans rely on correct AI advice and the decision was initially not correct, i.e., in which humans rightfully change their mind to follow the correct advice.

\begin{equation}
    Relative\;AI\;reliance\;(RAIR) =  \frac{\sum_{i=0}^{N}CAIR_i}{\sum_{i=0}^{N}CA_i}
\end{equation}

$CAIR_i$ is one if, in this particular case, the original human decision was wrong, the AI recommendation was correct and the human decision after receiving the AI recommendation is correct, and zero otherwise. $CA_i$ is one if the original human decision was wrong and the AI advice was correct, regardless of the final human decision, and zero otherwise. 
On the second dimension, we propose to measure the relative amount of correct self-reliance in the presence of incorrect AI advice.

\begin{equation}
    Relative\;self{\text-}reliance\;(RSR) = \frac{\sum_{i=0}^{N}CSR_i}{\sum_{i=0}^{N} IA_i}
\end{equation}

$CSR_i$ is one if on this particular instance the initial human decision was correct, the AI advice was incorrect and the human decision after receiving AI advice is correct. $IA_i$ is one, if the initial human decision for a task instance $i$ was correct and the AI advice was incorrect.

\begin{figure}[h]
  \centering
  \includegraphics[width=0.45\linewidth]{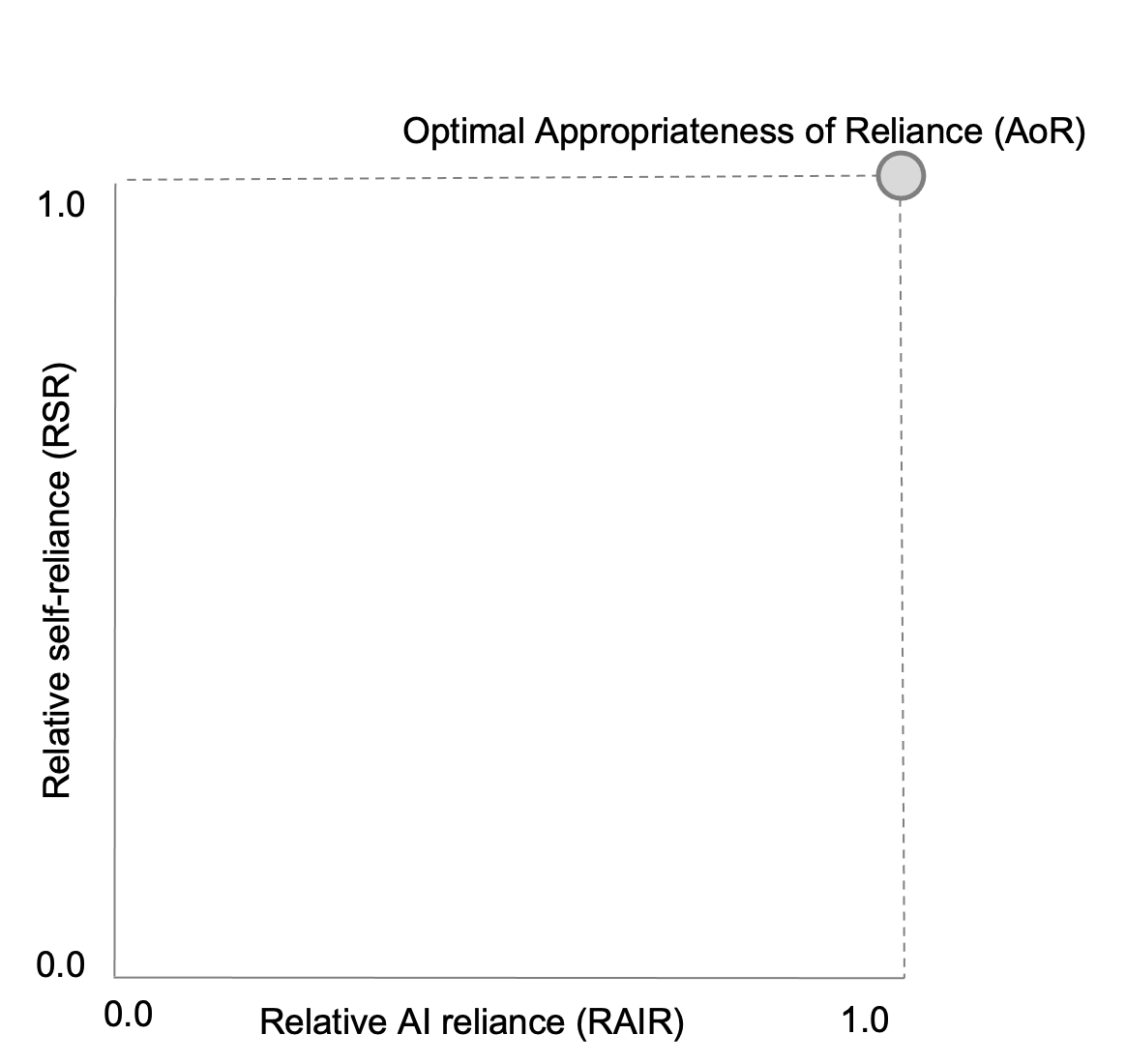}
  \caption{Two-dimensional depiction of the Appropriateness of Reliance (AoR).}
  \label{fig:dimension}
\end{figure}

\Cref{fig:dimension} on page \pageref{fig:dimension} highlights both dimensions. On the x-axis, we depict the relative  AI reliance (\(RAIR\)), and on the y-axis, the relative self-reliance (\(RSR\)). 
The figure highlights the properties of the measurement concept. It ranges between 0 and 1 along both dimensions. 
We call the tuple of $RAIR$ and $RSR$ \textit{Appropriateness of Reliance} (AoR).

\begin{equation}
    Appropriateness\;of\;Reliance\;(AoR) = (RSR; RAIR)
\end{equation}

We refer to the theoretical goal of having a \(RSR\) and a \(RAIR\) metric of ``1'' as \textit{optimal AoR}. 
Most likely, this theoretical goal will not be reached in any practical context as humans will not always be able to perfectly discriminate on a case-by-case basis whether they should rely on AI advice. Furthermore, random errors will reduce AoR as they cannot be discriminated against. Therefore, optimal AoR will most likely be a theoretical goal. 

\subsection{Definition of Appropriate Reliance}
So far, we have discussed how to measure AoR and the theoretical upper boundaries of AoR. The challenge, however, is to define the level of \(RAIR\) and \(RSR\)  that constitutes AR. In our work, we take an objective-oriented perspective and propose to define AR individually in the context of the task. Dependent on the task, different levels of \(RAIR\) and \(RSR\) might be appropriate.
In this work, we focus on AR with respect to performance ($P$) following \citet{talone2019effect,wang2021explanations}.\footnote{
Note that besides this performance perspective, appropriateness could also be discussed from an ethical perspective. Since in high-stake decision-making humans have an oversight responsibility \cite{schoeffer2022explanations}.} Thus, we define AR with regard to CTP:

\begin{equation}
    Appropriate \ Reliance (AR) =     
    \begin{cases}
    1 ,& \text{if } P_{H\&AI} > max (P_H, P_{AI})\\
    0,              & \text{otherwise}
\end{cases}
\end{equation}

With $P_{H\&AI}$ being the performance after receiving AI advice, $P_H$ the individual human performance and $P_{AI}$ the individual AI performance. Essentially, this means any tuple of $RAIR$ and $RSR$ that leads to CTP is considered AR.

In summary, we derived a metric (AoR) and defined AR. In the next section, we will use this foundation to derive a research model to analyze the impact of AI recommendation explanations on AoR.

\section{Theory Development And Hypotheses}
\label{sec:RM}
With the measurement concept (AoR) at hand, we now develop hypotheses on the effect of explanations on AoR.
Research has already investigated in-depth the effect of explanations on AI advice acceptance \cite{shin2021effects}. However, research is missing theoretical and empirical evidence on how explanations influence AoR \cite{bansal2022workshop}. Thus, our work contributes a research model that is evaluated applying the AoR concept.

As a dependent variable, we use the before-defined two dimensions of AoR--- namely $RAIR$ and $RSR$.
We believe that both dimensions need to be treated differently as there are inherent differences between the $RAIR$ and the $RSR$.
The $RAIR$ essentially deals with cases where the human is initially incorrect, gets correct advice, and relies on this advice. 
In contrast, $RSR$ focuses on cases where the human is initially correct and receives incorrect advice, which is then rightfully ignored. 
Thus, we formulate different hypotheses for the $RAIR$ and the $RSR$.

\begin{figure}[h]
  \centering
  \includegraphics[width=0.7\linewidth]{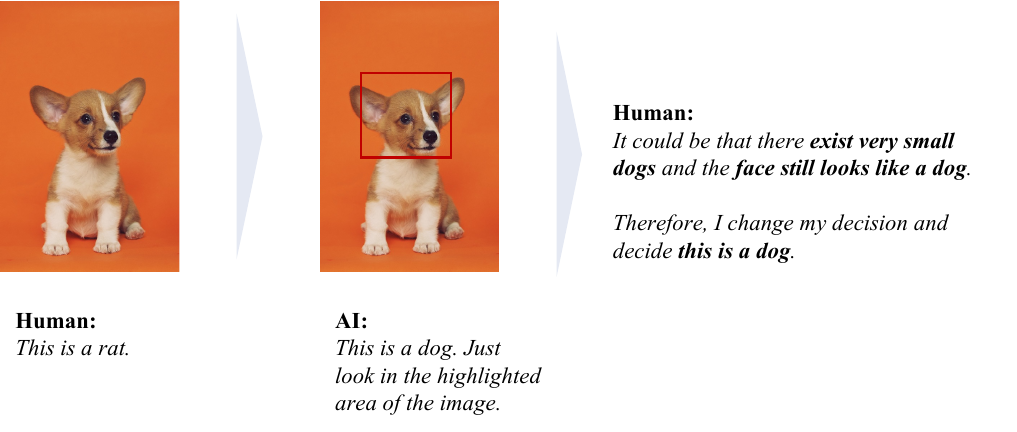}
  \caption{Illustrative example of how humans can increase their relative AI reliance ($RAIR$) based on explanations of the AI advice.}
  \label{fig:RAIR}
\end{figure}

The most central impact factor in the assessment of the AI advisor might be the explanations of its recommendations.
These explanations should provide insights into the AI's thought process. 
In the presence of incorrect advice, explanations might enable the human to detect whether the advice is incorrect, for example, by validating if the AI advisor violates some universal axioms of the task. Similarly, \cite{bansal2021does} hypothesize that if explanations do not ``make sense'', humans will reject the AI advice. 
However, on the other hand, sometimes explanations are rather interpreted as a general sign of competence and thereby increase over-reliance \cite{buccinca2021trust}. Which effect exceeds the other is unclear. Therefore, we hypothesize, without specifying a direction of the effect, that the explanations have a general effect on the $RSR$:

\textbf{H1a:} {\itshape Providing explanations of the AI advisor influences the relative self-reliance ($RSR$).}\\
The second effect of explanations on AoR is through the $RAIR$. Essentially, the $RAIR$ measures the percentage of times decision-makers follow the correct advice after initially being wrong about the task instance. This means they do not have enough domain knowledge to solve the task on their own. Thus, to increase the $RAIR$, human decision-makers need to extend their knowledge and simultaneously validate whether this knowledge extension makes sense.
Here, explanations are needed to first get inspired to derive new knowledge and second to validate the knowledge.
\Cref{fig:RAIR} shows an illustrative example based on an animal classification task. Imagine that a child has just seen big dogs and then sees a very small dog. It might think that this animal is something else, like a rat. Next, the child receives AI advice that says the animal it sees is a dog, and provides additional justification by highlighting the part of the image that led to the AI's decision. Now, the child's first task is to figure out whether this advice makes sense in general, while building the knowledge base. In this illustrative case, it might understand that the animal has characteristics of a dog, but is only smaller and therefore relies on the AI, thereby increasing the $RAIR$.
In the presence of correct advice, explanations might point humans towards new patterns they have not seen before and help discriminate these knowledge extensions.
In the presence of correct advice, the convincing element of explanations would not have a negative effect as a higher overall reliance on the AI advice would simply increase the $RAIR$.
We therefore hypothesize:

\textbf{H1b:} {\itshape Providing explanations of the AI advisor increases the relative  AI reliance ($RAIR$).}

Beyond the provisioning of additional information, explanations might change the attitude toward the AI advisor. 
In 1992, \citet{lee1992trust}, already discussed the influence of self-confidence and trust as predominant attitudes for reliance decisions in the context of automation.
Therefore, in the following, we discuss potential impacts on trust and the change in self-confidence induced through explanations and their impact on AoR.

Confidence is defined as a person’s degree of belief that their own decision is correct \cite{peterson1988confidence,zarnoth1997social}.
Confidence in one’s own decision is a key mechanism underlying advice acceptance \cite{wang2018does,chong2022human}.
So far, most research has discussed the influence of static human self-confidence on AoR, i.e. a confidence in doing the task instance without any advisor \cite{chong2022human}. 
However, we hypothesize that the absolute level of human confidence actually plays a minor role in comparison to the change of confidence after seeing the AI advice as it essentially reflects the combination of a self-assessment and the AI advisor's assessment. 
Explanations should not influence the initial human self-confidence but the confidence level after seeing the AI advice. 
Thus, we hypothesize:

\textbf{H2} {\itshape Providing explanations of the AI advisors increases the change in self-confidence.}\\
We hypothesize that this change in self-confidence positively correlates with the discrimination capability and thus should increase the $RSR$ and $RAIR$.

\textbf{H3a} {\itshape An increased change in human self-confidence increases the relative self-reliance (RSR).}

\begin{figure}[h]
  \centering
  \includegraphics[width=0.9\linewidth]{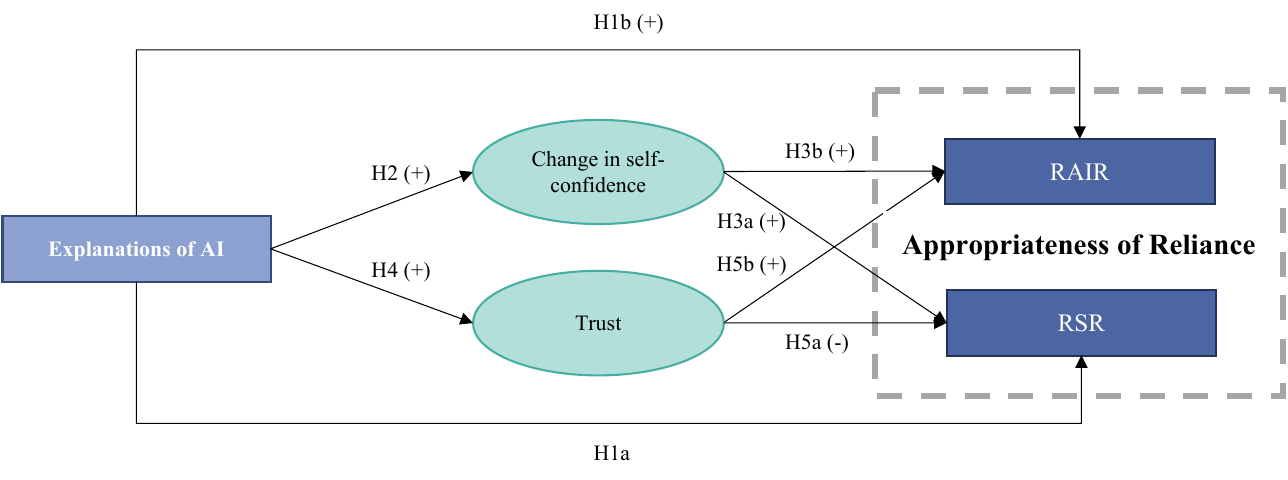}
  \caption{Research model on the effect of explanations on Appropriateness of Reliance (AoR).}
  \label{fig:SEM}
\end{figure}

\textbf{H3b:} {\itshape An increased change in human self-confidence increases the relative AI reliance (RAIR).}\\
We hypothesize that also trust in the AI advisor influences AoR. 
There are different levels of trust, e.g. trust in AI in general, trust in a specific AI advisor \cite{jacovi2021formalizing} and some researchers even refer to the case-by-case discrimination as trust on a task instance level \cite{wang2021explanations}. In this work, we focus on the specific trust in our developed AI advisor. 
In general, trust is a complex, multidisciplinary construct with roots in diverse fields such as psychology, management and information systems \cite{niese2022can}.
In our study, we define trust as a belief in the integrity, benevolence, trustworthiness, and predictability of the AI advisor following \cite{crosby1990relationship,doney1997examination,ganesan1994determinants,mcknight2002impact}.
Understanding is crucial in building trust \cite{gilpin2019explaining}.
Psychological research shows that in general explanations of humans increase trust \cite{koehler1991explanation}. Thus, we hypothesize:

\textbf{H4} {\itshape Providing explanations of the AI advisor increases trust in the AI advisor.}\\
Trust influences reliance, but does not fully determine it \cite{lee2004trust}.
When people show a high level of trust in the advisor, they consider the advice to be high-quality advice from an advisor with good intentions, and they will give more weight to that advice \cite{wang2018does}. In general, this should increase the acceptance of AI consulting.
For systems’ most effective use, however, users must appropriately  trust AI advisors \cite{lee2004trust}. 
Trust should be calibrated and match the AI's capabilities \cite{lee2004trust}.  Insufficient trust is called distrust and when trust exceeds capability it is called over-trust  \cite{lee2004trust}.
Research on automation has shown that over-trust can result in over-reliance on automation \cite{parasuraman1993performance,goddard2012automation,bailey2007automation} and, therefore, should decrease the $RSR$.
Thus, we hypothesize:

\textbf{H5a:} {\itshape Trust decreases the relative self-reliance (RSR). }\\
Since trust increases reliance it should also increase the $RAIR$. Therefore, we hypothesize: 

\textbf{H5b:} {\itshape Trust increases the relative AI reliance (RAIR). }\\
\Cref{fig:SEM} page \pageref{fig:SEM} highlights all our hypotheses and combines them into one integrated research model. 

\section{Experimental Design}
\label{sec:ED}
In this section, we present our study task, the AI model, and the corresponding explanations. We then explain the study procedure and measurements.

\subsection{Task, Model, and Explanations}
As an experimental task, we have chosen a deceptive hotel review classification. Humans have to differentiate whether a given hotel review is deceptive or genuine. 
\citet{ott2011finding,ott2013negative} provide the research community with a data set of 400 deceptive and 400 genuine hotel reviews. The deceptive ones were created by crowd-workers, resulting in corresponding ground truth labels.

\begin{figure}[h]
  \centering
  \includegraphics[width=0.7\linewidth]{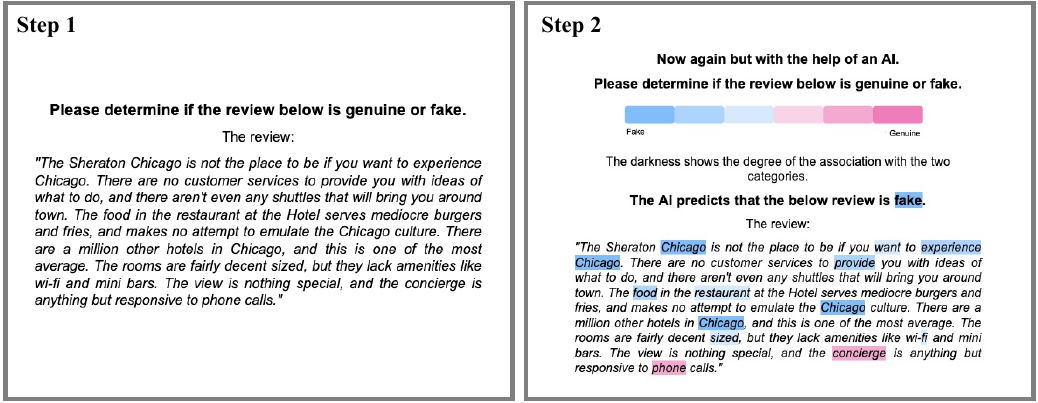}
  \caption{Online experiment graphical user interface for the feature importance condition. The ground truth of the exemplarily shown hotel review is ``fake''. The design of the interface is adapted from \citet{lai2020chicago}.}
  \label{fig:GUI}
\end{figure}

The implemented AI advisor is based on a Support Vector Machine with an accuracy of 86\%, which is a performance that is similar to the performance in related literature \cite{lai2020chicago}. For the explanations, we use a state-of-the-art explanation technique, LIME feature importance explanations \cite{ribeiro2016model}, as it is the most common one for textual data.
Feature importance aims to explain the influence of an independent variable on the AI's decision in the form of a numerical value. Since we deal with textual data, a common technique to display the values is to highlight the respective words according to their computed influence on the AI's decision \cite{lai2020chicago}. We additionally provide information on the direction of the effect and differentiate the values into three effect sizes following the implementation of \citet{lai2020chicago} (see step 2 in \Cref{fig:GUI}).

\subsection{Study Design and Procedure}
The research model is tested in an online experiment with a between-subject design.
We tested two different conditions. First, a \textit{control} condition in which the human receives AI advice without feature importance explanations and second, a \textit{feature importance} condition. 

In each condition, participants are provided with 16 reviews. We incorporate an advanced sampling strategy to isolate the effects of discriminating AI advice. We have a test set of 32 reviews to which we apply stratified sampling and select four reviews of each class of a confusion matrix (True Positive, False Positive, True Negative, False Negative), two with a positive sentiment and two with a negative sentiment. This approach allows us to ensure a high-performing AI that should provide good explanations but also the potential for incorrect AI advice. 

\textbf{Task flow.}
The online experiment is initiated with an attention control question that asks participants to state the color of grass. To control for internal validity, participants are randomly assigned to the condition groups. Then, both condition groups receive an introduction to the task and either AI alone or AI including feature importance. 
We provide the participants with a general intuition of the AI but not with specific performance information.
Then, the participants conduct two training tasks, to familiarize the participants with the task and the AI and, depending on the condition, with its explanations. Additionally, the participants receive feedback on the training tasks. After the two training reviews, the participants are provided with the 16 main tasks.  
For the AoR measurement concept, sequential task processing is essential. In our study, this means the human first receives a review without any AI advice, i.e., just plain text, and classifies whether the review is deceptive or genuine (see step 1 in \Cref{fig:GUI}). Then the participant is asked to classify the review and provide a confidence rating.
Following that, the human either receives a simple AI advice statement, e.g. ``the AI predicts that the review is fake'' or the AI advice and additional explanations (see step 2 in \Cref{fig:GUI}).
After receiving the AI advice the participant is able to change the initial decision and provide a new self-confidence assessment.
This sequential two-step decision-making allows us to measure AoR.
During the main tasks, the participants do not receive feedback on their performance. 
After classifying the hotel reviews, we collect data on trust and demographic variables.

\textbf{Reward.}
To incentivize the participants, they were informed that for every correct decision, they get an additional 12 Cents in addition to a base payment of 5.83 Euro. Hereby, the two training classifications do not count for the final evaluation. 

\textbf{Participant information.} 
The participants are recruited using the platform ``Prolific.co''.
We note that crowd workers might limit the generalizability of our results.
However, deception detection of digital information is often done in online communities. Future work could analyze the effects in professional deception detection screening services. 
In total, we conducted the experiment with 200 participants.
We excluded one participant in the feature importance condition because of a failed manipulation check. 
\Cref{tab:Participants} shows the age, gender, and education distribution of the participants.

\begin{table}[H]
\caption{Summary of Participants' Characteristics.}
\label{tab:Participants}
\begin{tabular}{ll}
\toprule
Number per condition & Control = 100   \\
& Feature importance = 99 \\
\midrule
Age & $\mu$ = 27.5, $\sigma$ = 8.5 \\
\midrule
Gender &  46 \% Male\\
 &  54 \% Female\\
 \midrule
 Education & 32 \% High school \\
 & 38 \% Bachelor  \\
 & 14 \% Master \\
 &  16 \% Other\\
\bottomrule
\end{tabular}
\end{table}

\subsection{Evaluation Measures}
To measure AoR, we use the upfront derived measurements $RAIR$ and $RSR$.
We measure the \textit{change in self-confidence} for all task instances per participant:

\begin{equation}
 Change\;in\;self{\text -}confidence = \sum_{i=1}^{16} Conf_{H2},i - Conf_H,i
\end{equation}

$Conf_H$ refers to the human self-confidence when doing the task instance alone and $Conf_{H2}$ to the human self-confidence after receiving AI advice. Both are measured with a 7-point Likert scale (``How confident do you feel in your decision?'').
We measure the trust in the AI advisor as a subjective latent construct with a 7-point Likert scale. We use four items based on \cite{crosby1990relationship,doney1997examination,gefen2003trust,ganesan1994determinants}. The items were: ``I think I can trust the AI.'', ``The AI can be trusted to provide reliable support.'', ``I trust the AI to keep my best interests in mind.'' and ``In my opinion, the AI is trustworthy.''. Cronbach's Alpha was 0.89 (high). The original scales were validated. 
As both classes are equally distributed, task performance was measured by the percentage of correctly classified images, i.e., accuracy. To measure the \textit{human accuracy}, we calculated this measure for both conditions based on the initial human decision across all 16 task instances. Furthermore, we calculate the \textit{AI-assisted accuracy} based on the revised human decision after receiving AI advice. 
\section{Results}
\label{sec:Results}
In the following, we present the results of our behavioral experiment. We start by presenting descriptive results, followed by the results with respect to AoR and AR. Following that, we analyze our full research model, including mediations, by applying Structural Equation Modelling (SEM).

\subsection{Descriptive analysis}
Descriptive results of our study can be found in \Cref{tab:descriptives}.
They are split according to the experimental condition. We evaluate the significance of the results using t-tests  after controlling for normality. 
The participants' $RAIR$ is significantly higher in the explanation group compared to the control group $(t =-1.95 , p = 0.05)$. 
The change in confidence is statistically significant $(t =-2.33 , p = 0.02)$ which means that on average people feel more confident after receiving AI advice including explanations. 
Neither AI-assisted nor human accuracy is significantly different between conditions. However, the difference between the human and AI-assisted performance is significant (feature importance condition = 2.45 pp; control condition = -1.56 pp ; $t = 2.29, p = 0.02$) which means that explanations not only improve the $RAIR$ but also as a consequence the overall performance.

\begin{table}[H]
\caption{Descriptive outcomes.}
\label{tab:descriptives}
\begin{tabular}{lccccccc}
\toprule
Condition  & $RAIR$ & $RSR$  & Trust (SD) & Change in Self- & AI-assisted & Human\\
 & &   &  &  Confidence (SD) & Accuracy & Accuracy \\
\midrule
Control  & 29.59 \%  & 71.87\%  & 4.45 (1.17) & 0.19 (0.42)  & 53.94 \%  & 55.50 \%   \\
 && &  & &  &  \\
Feature&  38.87\%  &  69.45 \%   & 4.4 (1.18)& 0.06 (0.35) & 56.30 \%  & 53.85 \%  \\
Importance  &  & &  &  &  &  \\

\bottomrule
\end{tabular}
\end{table}

\subsection{Appropriateness of Reliance \& Appropriate Reliance}
\begin{figure}[h]
  \centering
  \includegraphics[width=0.45\linewidth]{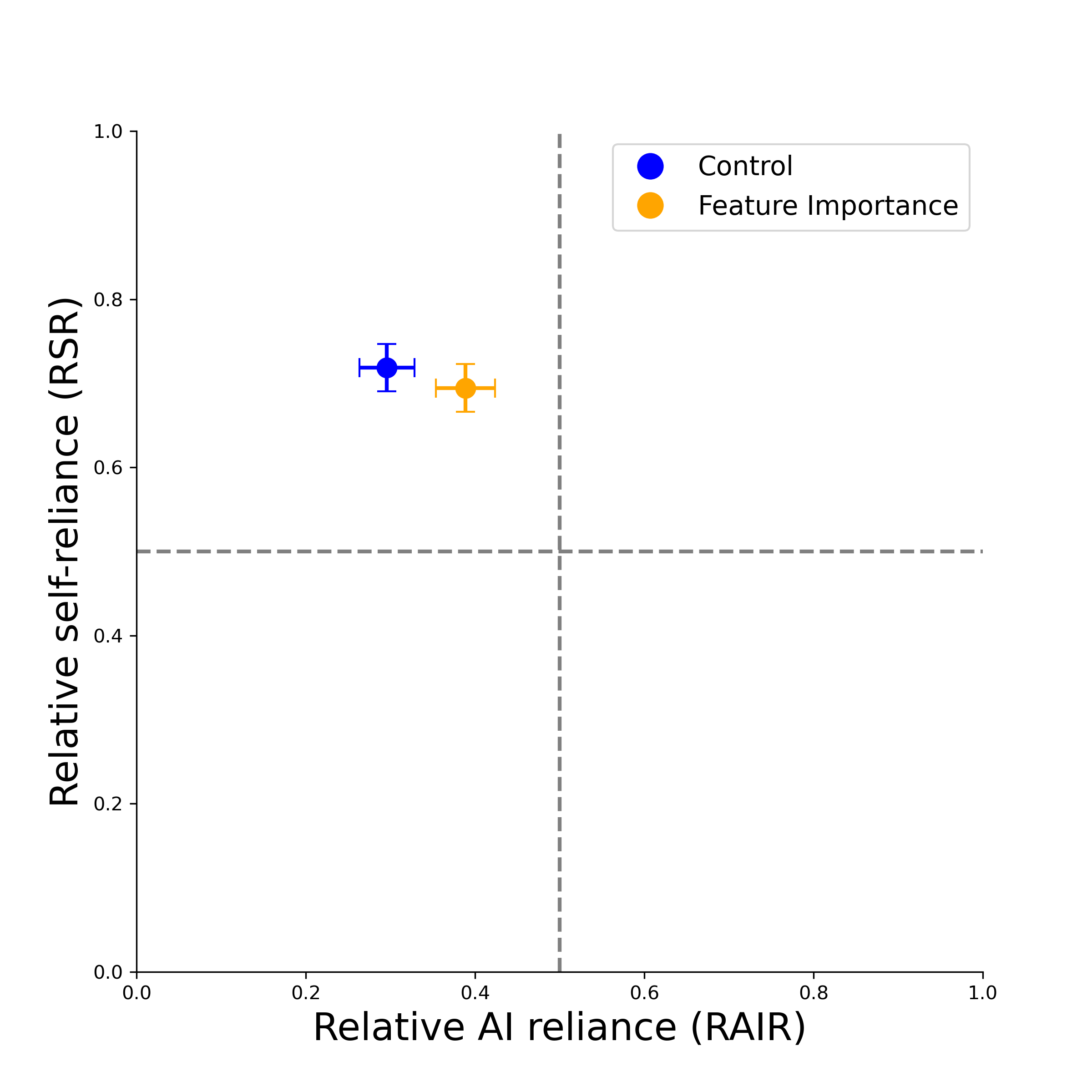}
  \caption{Illustration of Appropriateness of  Reliance (AoR) including standard errors. Explanations increase the $RAIR$ significantly. Differences in $RSR$ are not significant.}
  \label{fig:illustration}
\end{figure}

We depict our AoR results of the experiment in \Cref{fig:illustration}. They highlight in the control condition a high \(RSR\) of $71.87 \%$ ($\pm3 pp$) and a relatively low \(RAIR\) of $29.59 \%$ ($\pm 3 pp$). 
This indicates that humans in the setting were able to differentiate between wrong AI advice and self-rely to a high degree. The \(RAIR\) of $29.59 \%$ shows that we can observe a severe share of under-reliance on AI.

In the XAI condition, we can observe a significant increase ($t=-1.95$, $p =0.05$) in \(RAIR\) from $29.59 \%$ ($\pm3 pp$) to $38.87 \%$ ($\pm3 pp$) while the \(RSR\) does not change significantly ($71.87 \% \pm3 pp$ for the control condition and $69.45 \%\pm3 pp$ for the feature importance condition, $t = 0.61$, $p = 0.54$). This means explanations of AI decisions can reduce the share of under-reliance. It is important to highlight that $RAIR$ is not increased simply by relying more often on AI advice, as this would have also reduced the \(RSR\) significantly. Thus, our experiment indicates that feature importance on textual data can have a positive effect on human-AI decision-making. 

Following our AR definition, to evaluate whether the participants display AR, we need to calculate whether we reached CTP. Therefore, we compare the individual human and AI performance with the human-AI team performance. The down-sampled AI performance is 50 \% for both conditions. The human accuracy varies depending on the condition. The human-AI team performance is not significantly different from the human accuracy which means we do not reach CTP and therefore AR is not displayed. Further means would be necessary to reach AR.

\subsection{Structural Equation Model}

In addition to analyzing the direct effect of explanations on the $RAIR$ and $RSR$, we 
use SEM analysis to test our hypothesized research model.
Before fitting our SEM, we conducted  missing data analysis, outlier detection, a
test for normality, and the selection of an appropriate estimator. We observe no missing data and no outliers. However, one participant failed our attention check leaving us with a final sample size of 199 for both conditions.
Shapiro's test for normality indicates that several variables of interest deviate significantly from normal distributions.
As a result, we conducted the analysis with an estimator that
allows for robust standard errors and scaled test statistics \cite{kunkel2019let}. Therefore, we use
the MLR estimator \cite{lai2018estimating}.

\begin{table}[H]
\caption{Structural equation model fitting index using a chi-squared test ($\chi^2$), root mean square error of approximation (RMSEA), Comparative Fit Index (CFI), Tucker-Lewis Index (TLI) and Standardized Root Mean Squared Residual (SRMR).}
\label{tab:SEM_test}
\begin{tabular}{lccccc}
\toprule
 & $\chi^2$ & RMSEA & CFI & TLI  & SRMR\\
\midrule
Measurement criteria & > 0.05 \cite{bentler1990comparative} & < 0.05 \cite{hu1999cutoff}  & > 0.96 \cite{hu1999cutoff} & > 0.95 \cite{hu1999cutoff} & < 0.08 \cite{hu1999cutoff} \\
Value & 0.083 & 0 & 1 & 1.02 & 0.02\\
\bottomrule
\end{tabular}
\end{table}

\begin{figure}[h]
  \centering
  \includegraphics[width=0.9\linewidth]{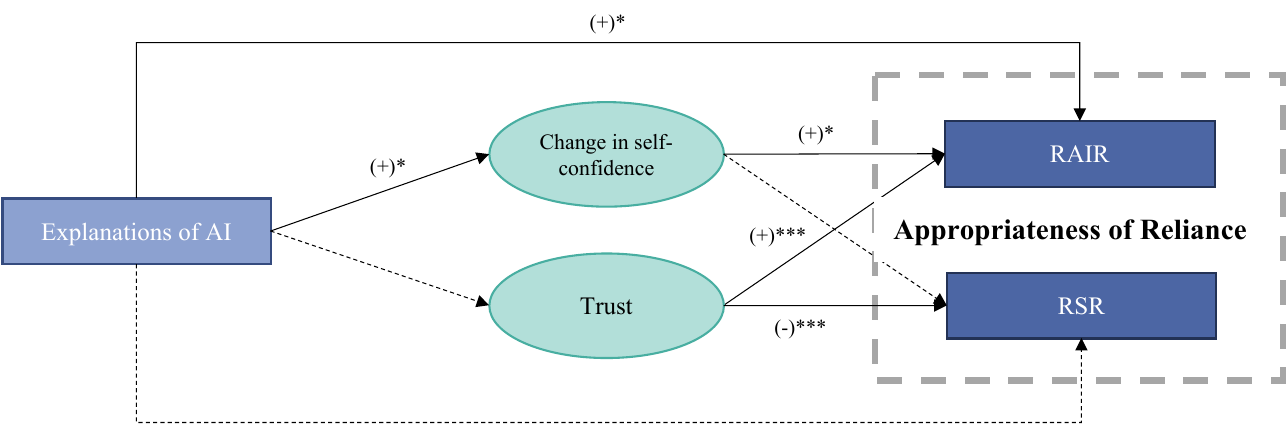}
  \caption{
  Structural equation modeling results.
  Significance: $\star\star\star$ p < 0.01, $\star\star$ p <0.05, $\star$ p < 0.1. 
  }
  \label{fig:experiment}
\end{figure}

Our dependent variables are the $RAIR$ and $RSR$. Since
these dependent variables are between 0 and 1, we employed a logistic model in the Lavaan package, version 0.6-9, in R \cite{lavaan}. 
This model has an excellent overall fit (see \Cref{tab:SEM_test}). 
The results for each independent
variable are discussed below and visualized in \Cref{fig:experiment}. 

We do not find any significant relationship between providing explanations of the AI advisor and the $RSR$ (\textbf{H1a}).
However, also in the SEM the effect of explanations on the $RAIR$ is significant (\textbf{H1b}).
Additionally, we find a significant positive relationship between providing explanations of the AI advisor and the change in self-confidence which confirms \textbf{H2}. We also observe a positive correlation between the change in self-confidence and the $RAIR$ (\textbf{H3b}). Since the absolute strength of the relationship also decreased with including the change in confidence, we can conclude that the positive effect of explanations on the $RAIR$ is partially mediated by the change in confidence. 
Interestingly, we do not find a relationship between the change in self-confidence and the $RSR$ (\textbf{H3a}).
Additionally, we find no effect of our explanations on trust (\textbf{H4}). However, both \textbf{H5a} and \textbf{H5b} are confirmed, i.e. increasing trust increases $RAIR$ but also decreases $RSR$.
We display a summary of the hypotheses results in \Cref{tab:SEM}.
In the following Section, we discuss our results.

\begin{table}[H]
\caption{Analysis results of the structural equation model ($\star\star\star$ p < .01, $\star\star$ p <.05, $\star$ p < .1)}
\label{tab:SEM}
\begin{tabular}{llcll}
\toprule
X & Y & z-value & Standardized regression  & Result \\
 &  & & coefficient &  \\
\midrule
H1a: Explanations --> & $RSR$ &  0.04 & -0.03  & not supported \\
H1b: Explanations --> & $RAIR$ &  1.73 & 0.08 $\star$ & supported  \\
H2: Explanations --> & Change in self- confidence &  2.30 & 0.13 $\star$ & supported \\
H3a: Change in confidence  --> & $RSR$ &  0.02 & 0.00 & not supported \\
H3b: Change in confidence  --> & $RAIR$ &  1.84 & 0.11 $\star$ & supported \\
H4: Explanations  --> & Trust &  -0.21 & -0.04 & not supported \\
H5a: Trust  --> & $RSR$ &  -3.08 & -0.05 $\star\star\star$ & supported \\
H5b: Trust  --> & $RAIR$ &  3.19 & 0.06 $\star\star\star$ & supported \\
\bottomrule
\end{tabular}
\end{table}



\section{Discussion}
\label{sec:Discussion}
In this article, we first defined AR and conceptualized a measurement concept (AoR). Following that, we derived a research model regarding the impact of explanations on AoR which we subsequently tested on a deception detection task. 
Though conducted in a limited scope, our findings should help to guide future work on AR.

\textbf{Theoretical foundation of appropriate reliance.}
The main contribution of our work is the theoretical development of AR. 
So far, terms like ``appropriate trust'', ``calibrated trust'' and AR were often used interchangeably in prior research. We provide clarity by defining AR and putting the terms in perspective.
Second, we derive a granular, two-dimensional measurement concept---Appropriateness of Reliance (AoR). Most prior work neglected the initial human decision that would have been made without any AI advice.\footnote{The one exception is the work of \cite{buccinca2021trust} who measure over-reliance in the same way that we do but do not consider under-reliance.} Taking this initial human decision into account allows to differentiate between the effects of advice (correct and incorrect) and confirmation. Without the human initial decision, we cannot say for sure whether a final wrong human decision is due to over-reliance on a wrong AI advice or due to the human and the AI being wrong.

\textbf{Implications for appropriate reliance and explainable AI.}
We further have investigated the effect of AI explanations on AoR.
We confirm the results of prior research \cite{bansal2021does,gonzalez2020human,wang2021explanations} by finding an effect of explanations on the $RAIR$.
We believe that the reason for this could be that the AI's explanations increase people's knowledge of the task \cite{gonzalez2020human,spitzer2022training}. 
Maybe in such cases, the human should be seen less as a ``judge'' but instead more as a student of the AI. 
On the other hand, our results show that explanations do not influence the $RSR$. While this may sound disappointing at first, it also shows that the claim that explanations would reduce overreliance \cite{bansal2021does,buccinca2021trust} does not seem to hold for all kinds of tasks.\footnote{An earlier study by this research team \cite{schemmer2022influence} found initial signs of a reduced $RSR$ in a pretest. However, this study with more participants shows that the effect is not significant in a larger sample.}
It further suggests that new techniques must be developed to distinguish incorrect AI advice.
Moreover, our study is the first one that analyzed mediators of the effect of explanations on AR. 
Interestingly, in our study, we find no effect of explanations on trust. However, prior research has shown that it depends on a lot of confounding factors. 
We find significant effects of trust on $RAIR$ as well as $RSR$. 
Additionally, we show that the effect of explanations on $RAIR$ partially depends on the change in confidence after receiving AI advice. 

\textbf{Appropriate reliance and complementary team performance (CTP).}  
Lastly, we want to elaborate on the relationship between AoR and CTP. To reach CTP, the task needs to have instances where the AI is better than the human and vice versa, i.e. a certain amount of complementarity potential needs to be present \cite{hemmer2022effect}. 
CTP essentially depends on the relationship between $RSR$ and $RAIR$ and this complementarity potential.

The human impact on CTP is given by multiplying the $RSR$ and the share of incorrect AI advice. However, simultaneously involving a human might reduce the AI performance ($1 - RAIR$ multiplied by the share of correct advice).
That means to reach CTP, the gain through human involvement needs to be larger than the loss through discounting correct AI advice or more formally\footnote{Provided that nothing changes in the cases where the initial decision of the human and the AI advice are the same.}:

\begin{equation}
    CTP = 
    \begin{cases}
    1 ,& \text{if } RSR * IA > (1- RAIR) * CA\\
    0,              & \text{otherwise}
\end{cases}
\end{equation}

Here $IA$ is the total number of task instances in a test set where the human is initially correct and receives incorrect advice.
$CA$ refers to the number of task instances where the human is initially incorrect and receives correct advice.
If this condition is not fulfilled, AR depends on the relationship between human and AI performance. 
If the human performs worse than the AI advisor and has a low $RSR$ and $RAIR$, one should always favor AI advice. If the human decision-maker performs better on average than the AI, one can argue that from a performance perspective the AI should not be used.

\textbf{Limitations.}
No research is without limitations.
We would like to emphasize that deception detection is a difficult task for humans \cite{lai2019human,lai2020chicago}. Humans on average just perform slightly better than by chance on this particular hotel review task \cite{lai2020chicago}. This makes AR difficult as the task of discrimination requires human domain knowledge. 
Additionally, the generalizability of our experimental findings is limited due to the choice of explanations. However, we have deliberately chosen the most modern form of explanation to maximize the impact of our results. 
Our concept is limited to classification tasks but will be extended in future work. First approaches can be found in the work of \citet{petropoulos2016big}.

Furthermore, the sequential task setup necessary for our measurement concept has some disadvantages as it changes the task itself. Since conducting the same task initially alone before receiving AI advice, the human is already mentally prepared and might react differently than after directly receiving AI advice. More specifically, research has shown that letting humans conduct the task alone before receiving AI advice might reduce over-reliance \cite{buccinca2021trust}. The sequential task setup could induce an anchoring effect which prevents the human to more actively take the AI into account \cite{buccinca2021trust}. This could have led to the overall low $RAIR$ in our experiment.
Moreover, sequentially conducted tasks with AI advice might not always be possible or desired in real-world settings. Therefore, the measurement should be seen as an approximation of real human behavior. 
Instead of having a sequential task setup, one alternative option could be to simulate a human model based on a data set of task instances solved by humans without AI advice. This simulation model could approximate the initial human decision within a non-sequential task setting. However, this approach is also an approximation of real human behavior. 
In other work, a latent construct has been derived to measure reliance behaviour \cite{tejeda2022ai}. 
Future work should compare the approaches.

\textbf{Future Work.}
Essentially, the improvement in $RAIR$ depends on the knowledge gain of the human.
Future work could therefore extend our research model by adding newly learned knowledge as a mediator. Empirically, this could be measured by asking humans before collaborating with AI to do a couple of task instances on their own and afterwards \cite{spitzer2022training}. The performance improvement can be interpreted as learned knowledge.

Most importantly, future research needs to investigate the impact of different design features of AR. We initiate our research with state-of-the art feature importance but many other ones can be thought of, e.g. counterfactuals, global explanations. Future research should evaluate these potential design features to provide practitioners with a toolkit for effective use of AI.
\section{Conclusion}
\label{sec:Conclusion}
Appropriate reliance in AI advice is the next milestone after a decade of research focused on AI adoption and acceptance.
Nowadays, many AI applications are deployed and used on a daily basis. While adoption and acceptance remain important, we argue that a perspective shift is necessary. In the use phase of AI, researchers need to find ways to ensure appropriate reliance and, thus,  effective
use of AI.  In this article, we provide guidance for future research on appropriate reliance by providing a definition and a
measurement concept---Appropriateness of Reliance (AoR). Furthermore, we generate initial insights how explanations influence the Appropriateness of Reliance. We hope that our research will inspire researchers and practitioners
for future research on appropriate reliance, resulting in effective human-AI collaboration.

\bibliographystyle{ACM-Reference-Format}
\bibliography{99_bib}

\end{document}